\documentclass[a4paper]{article}
\usepackage{INTERSPEECH2019}

\usepackage{amsmath,graphicx, enumitem, multirow, subcaption, color}
\usepackage{url}

\let\svthefootnote\thefootnote

\title{On the Choice of Modeling Unit for Sequence-to-Sequence Speech Recognition}
\name{Kazuki Irie$^{1*}$, Rohit Prabhavalkar$^2$, Anjuli Kannan$^2$, Antoine Bruguier$^2$,\\ David Rybach$^2$, Patrick Nguyen$^2$}
\address{$^1$Human Language Technology and Pattern Recognition Group, Computer Science Department \\
         RWTH Aachen University, D-52056 Aachen, Germany\\
         $^2$Google, Mountain View, CA 94043, USA}
\email{irie@cs.rwth-aachen.de, \{prabhavalkar, anjuli, tonybruguier, rybach, drpng\}@google.com}
\begin{document}
%
\maketitle
\begin{abstract}
In conventional speech recognition, phoneme-based models
outperform grapheme-based models for non-phonetic languages such as English.
The performance gap between the two typically reduces as the amount of training data
is increased.
In this work, we examine the impact of the choice of modeling unit for
attention-based encoder-decoder models.
We conduct experiments on the LibriSpeech 100hr, 460hr, and 960hr tasks, using
various target units (phoneme, grapheme, and word-piece);
across all tasks, we find that grapheme or word-piece models consistently
outperform phoneme-based models, even though they are evaluated without a
lexicon or an external language model.
We also investigate model complementarity: we find that we can improve WERs by up to 9\% relative by
rescoring N-best lists generated from a strong word-piece based baseline with either the
phoneme or the grapheme model.
Rescoring an N-best list generated by the phonemic system, however, provides
limited improvements.
Further analysis shows that the word-piece-based models produce more diverse
N-best hypotheses, and thus lower oracle WERs, than phonemic models.
\end{abstract}
\noindent{\bf Index Terms}:
End-to-end speech recognition, word-pieces, graphemes, phonemes, sequence-to-sequence\let\thefootnote\relax\footnote{*Work performed during internship at Google.
We thank Tara Sainath and Yu Zhang for helpful discussion,
and Jinxi Guo for sharing his language model setup.
An initial version of this paper appears as a pre-print \cite{irie2019}.}
\addtocounter{footnote}{-1}\let\thefootnote\svthefootnote
\vspace{-2mm}
\section{Introduction}
\label{sec:intro}
Sequence-to-sequence learning~\cite{seq2seq} based on encoder-decoder attention
models~\cite{bahdanau2014neural} has become popular for both machine
translation~\cite{wu2016google} and speech recognition~\cite{RPis2017,
KimHW17, battenberg2017exploring, WengCWWYSY18, ChiuSWPNCKWRGJL18}.
Such models are typically trained to output \emph{character-based} units:
graphemes, byte-pair encodings (BPEs)~\cite{sennrich16bpe}, or
word-pieces~\cite{SchusterN12}, which allow the model to directly map
the frame-level input audio features to the
output word sequence, without using a hand-crafted pronunciation lexicon.
Thus, when using such character-based output units, end-to-end speech
recognition models~\cite{lasicassp2016} jointly learn the acoustic model,
pronunciation model, and language model within a single neural network.
In fact, such models outperform conventional hybrid recognizers~\cite{Bourlard1993} when trained on
sufficiently large amounts of data~\cite{ChiuSWPNCKWRGJL18}.

One of the main advantages of character-based sequence-to-sequence models lies in their simplicity:
both for training, as well as decoding. 
In fact, the use of characters as units for acoustic modeling has a long history for conventional
HMM-based automatic speech recognition (ASR) systems (e.g., \cite{kanthak02, KillerSS03, SungHBS09}, inter alia).
In the context of conventional ASR systems, for non-phonetic languages such as English, where the correspondence between orthography and pronunciation is less clear, previous works \cite{kanthak02, KillerSS03} have found that phoneme-based models outperform grapheme-based models; grapheme-based systems approach the performance of phoneme-based systems only when much larger amounts of training training data are available \cite{SungHBS09}.
It is therefore, natural to ask whether similar observations also apply to recently proposed attention-based encoder-decoder models: specifically, how do attention-based encoder-decoder models perform when using phonemes instead of character-based output units? 
To the best of our knowledge, this question has only been empirically investigated in the setting where a large amount of labeled training data are available. In previous work \cite{sainath2017no, zhou2018comparison}, it has been empirically shown that the grapheme-based encoder-decoder models outperform the phoneme-based approach, while \cite{sainath2017no} find that use of lexica is still useful for recognizing rare words such as named entities. 

In this work, we first investigate whether the previous result \cite{sainath2017no} which establishes the dominance of lexicon-free graphemic models over the phoneme-based models also hold on tasks with smaller amounts of training data.
We carry out evaluations on the three subsets of the LibriSpeech task \cite{panayotov2015librispeech}: 100hr, 460hr, and 960hr, where we find that grapheme or word-piece models do indeed consistently outperform phoneme-based models, even when training data is limited.
In Sec.~\ref{sec:part2}, we further investigate the benefits offered by phonemic models by studying the complementarity of different units. In experimental evaluations, we find that simple N-best list rescoring results in large improvements in WER. Finally, we conduct a detailed analysis of the differences in the hypotheses produced by the models with various output units, in terms of quality of the top hypotheses, as well as the oracle error rate of the N-best list.
\begin{figure}[b]
	\vspace{-5mm}
	\centerline{\includegraphics[width=0.4\columnwidth]{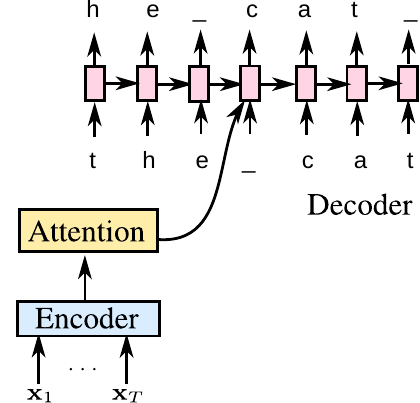}}
	\caption{\it LAS model.}
	\label{fig:las_base}
\end{figure}
\vspace{-2mm}
\section{Sequence-to-Sequence Speech Models}
\label{sec:model}
All our models are Listen, Attend, and Spell (LAS)~\cite{lasicassp2016} speech models.
The LAS model, which is depicted in Figure \ref{fig:las_base}, has encoder, attention, and decoder modules. 
The \textit{encoder} transforms the input frame-level audio feature sequence into
a sequence of hidden activations. The \textit{attention module} summarizes the
encoder sequence into a single vector for each prediction step, and finally,
the \textit{decoder} models the distribution of the output sequence conditioned
on the history of previously predicted labels.
Both the encoder and the decoder are modeled using recurrent neural networks,
and thus the entire model can be jointly optimized.
We refer the interested reader to~\cite{lasicassp2016, RPis2017, WeissCJWC17} for more details.
Standard LAS models use \textit{character-based} output units:
grapheme \cite{lasicassp2016}, word-piece \cite{ChiuSWPNCKWRGJL18} or BPE \cite{zeyer2018:asr-attention}.
\vspace{-2mm}
\section{Phonemic Sequence-to-Sequence Model}
\vspace{-1mm}
The phonemic LAS model can be obtained by using phonemes as the output unit.
Phonemes are natural labels for acoustic modeling of non-phonetic languages.
The use of a pronunciation lexicon can also ease integration of
completely new words or named entities\footnote{This might not be as relevant
for LibriSpeech evaluation as we use the official LibriSpeech lexicon without
modification.}\cite{bruguier19}.
However, by using a pronunciation lexicon, we give
up the end-to-end approach, which introduces complications for both training and
decoding.

For training, words with multiple pronunciation variants cause a problem, since
there is no unique mapping from such a word to its corresponding phoneme
sequence.
While we can potentially obtain the correct pronunciation variant by generating
alignments, we skip this extra effort by choosing a pronunciation simply by
randomly choosing one of the pronunciations for each word to define a unique
mapping.
In addition, we include an unknown token \texttt{UNK} as a part of the phoneme vocabulary and
use it to represent words which are not included in the lexicon.
We use a dedicated end-of-word token \texttt{EOW} (as part of the phoneme
inventory) to model word boundaries, as in~\cite{sainath2017no}, which we
find improves performance.

To deal with the ambiguity of homophones\footnote{By choosing phonemes as output units,
we are giving up the standalone recognition using the attention-based model. Also, while 
acoustic modeling motivates the use of phonemes, the ability
of the decoder as a language model can possibly be weaker compared with character-based units,
since the phoneme-level language model can be viewed as a subword-level class-based language model \cite{Brown92}
where the clusters are formed based on the phonemic similarity.}
during decoding, we incorporate a
(word-based) n-gram language model. We use a general weighted finite-state
transducer (WFST) decoder to perform a beam search. The lexicon and language
model (LM) are represented as WFST $L$ and $G$ respectively and combined by means of
FST composition as the search network $L\circ G$~\cite{mohri2008}. The search
process then explores partial path hypotheses which are constrained by the
search network and scored by both the LAS model and the n-gram language model.
\vspace{-6mm}
\section{LibriSpeech Experimental Setup}
 \subsection{Dataset}
 \vspace{-1mm}
The LibriSpeech task \cite{panayotov2015librispeech} has three subsets with different amounts of
transcribed training data: 100hr, 460hr, and 960hr. A lexicon with pronunciations for 200K words
is officially distributed. The development and test data are both split into \textit{clean}
and \textit{other} subsets, each of them consisting of about 5 to 6 hours of audio. The number of unique words
observed in each subset as well as the out-of-vocabulary (OOV; unseen in training
data) rate is summarized in Table~\ref{tab:oov}.
For language modeling, extra text-only data of about 800M words is also
available, along with an officially distributed 3-gram word LM; we use the
unpruned 3-gram LM for decoding the phonemic LAS models.
In contrast, the grapheme and word-piece models are evaluated without a lexicon or a language model (unless otherwise indicated).
We train word piece models~\cite{SchusterN12} of size 16K (16,384) on each training subset.
\begin{table}[h]
	\centering
	\caption{\it Out-of-vocabulary (OOV) rates (\%) with respect to the
	vocabulary (unique word list) in different data scenarios, and with respect to the pronunciation lexicon.}
	 \label{tab:oov}
	 	\vspace{-2mm}
	\begin{tabular}{ |c|c|c|c|c|c|} \hline
		Training  & Vocab.    & \multicolumn{2}{|c|}{dev} & \multicolumn{2}{|c|}{test}    \\ \cline{3-6}
		data (h)  & Size      & clean & other & clean & other    \\ \hline
		100   &  34~K &  2.5  & 2.5 & 2.4 & 2.8  \\
		460   &  66~K &  0.9  & 1.2 & 1.0 & 1.3  \\
		960   &  89~K &  0.6  & 0.8 & 0.6 & 0.8  \\  \hline \hline
		Lexicon  & 200~K   & 0.3  & 0.6 & 0.4 & 0.5  \\ \hline
	\end{tabular}
\vspace{-5mm}
\end{table}
\vspace{-2mm}
\subsection{Models and training}
\vspace{-1mm}
\label{sec:model_tr}
We use 80-dimensional log-mel features with deltas and accelerations as the
frame-level audio input features.
Reducing input frame rate in the encoder is important for successfully training
sequence-to-sequence speech models, especially for tasks such as LibriSpeech
which feature long utterances ($\sim$15s).
Thus, following~\cite{zhangCJ17}, our encoder layers include two layers of 3$\times$3 convolution
with 32 channels with a stride of 2, which results in a total time reduction factor of 4.
We consider three model (\textit{small}, \textit{medium}, and \textit{large})
which differ in terms of the sizes of model components.
On top of the convolutional layers, the encoder contains 3 (\textit{small}) or 4
(\textit{medium} and \textit{large}) layers of bi-directional
LSTMs~\cite{schuster1997bidirectional}, with either 256 (\textit{small}), 512
(\textit{medium}), or 1024 (\textit{large}) LSTM \cite{hochreiter1997long} cells
in each layer.
A projection layer and batch normalization are applied after each LSTM encoder
layer~\cite{zhangCJ17}.
The decoder consists of 1 (\textit{small}) or 2 (\textit{medium} and
\textit{large}) LSTM layers, and uses \emph{additive attention} as described in~\cite{WeissCJWC17}.

We train all models using 16 GPUs by asynchronous stochastic gradient descent
with Adam optimizer~\cite{kingma15} from random initialization without any
special pre-training method\footnote{We find training to be stable across repeated runs.
We avoid plateaus at the beginning of training (which we often observe) by tuning the initial learning rate.
We find that our models achieve the best WER on the dev-clean, earlier than on the dev-other set.} for about 80 epochs.
We use open-source Tensorflow Lingvo toolkit \cite{shen2019lingvo} for all experiments. Our
grapheme and word-piece based baseline configurations are publicly available online\footnote{\url{https://github.com/tensorflow/lingvo}}
where further details about the models can be found.

\vspace{-2mm}
\section{Standalone Performance Results}
\label{sec:part1}
\subsection{Baseline model performance on 960hr}
\vspace{-1mm}
The WER performance of grapheme and word-piece based models is summarized in
Table~\ref{baseline960}.
For both graphemes and word-pieces, we present the performance for \textit{small}, \textit{medium} and \textit{large}
model sizes (as shown by different numbers of parameters) as described in Sec \ref{sec:model_tr}. The difference of number of parameters between different units only comes from the unit-level vocabulary size.
As can be seen in Table \ref{baseline960}, models benefit from the additional parameters and the best
WERs are obtained for the large word-piece model.
\vspace{-2mm}
	\begin{table}[h]
		\centering
		\caption{\it WERs (\%) for grapheme and word-piece models.}
			 	\vspace{-2mm}
		\label{baseline960}
		\begin{tabular}{ |l|c|c|c|c|c|} \hline
			\multirow{2}{*}{Unit}      & \multirow{2}{*}{Param.}  & \multicolumn{2}{|c|}{dev} & \multicolumn{2}{|c|}{test}    \\ \cline{3-6}
			&   & clean & other & clean & other    \\ \hline
			\multirow{3}{*}{Grapheme}   & 7~M    & 7.6 & 20.5 & 7.9 & 21.3  \\
			& 35~M   & 5.3 & 15.6 & 5.6 & 15.8  \\
			& 130~M  & 5.3 & 15.2 & 5.5 & 15.3  \\ \hline
			\multirow{3}{*}{Word-Piece} & 20~M   & 5.8 & 16.0 & 6.1 & 16.4 \\
			& 60~M   & 4.9 & 14.0 & 5.0 & 14.1 \\
			& 180~M  & \textbf{4.4} & \textbf{13.2} & \textbf{4.7} & \textbf{13.4} \\ \hline
		\end{tabular}
	\end{table}
\subsection{Phonemic model performance on 960hr}
		\vspace{-1mm}
For phoneme based models, we first check the phoneme error rates (PER) in order to make
sure that the models are reasonable\footnote{By increasing the model size from 7~M to 35~M, then to 130~M, we improve
the PERs (\%) from (3.2, 9.7, 3.2, 9.9), to (2.8, 8.9, 3.0, 9.1), then to  (2.4, 7.9, 2.5, 7.7) on the 
dev-clean/other, test-clean/other sets.}.
The WER performance results for decoding with the
lexicon and the 3-gram word LM (88M n-grams) is shown in Table \ref{960overview}.
We observe that despite the use of an external LM
which is trained on much more data than the transcribed acoustic training data, the phonemic system performs worse
than the best graphemic model\footnote{This is similar to what is reported in \cite{sainath2017no}. Though, we note that we get about 2\% absolute degradation in WERs
with a model trained without \texttt{EOW} compared with the model with \texttt{EOW}.}.
It is nevertheless interesting to examine examples where the phonemic model outperforms the best word-piece
model. In Table \ref{plg_vs_wp}, we present some illustrative examples.
In addition, we find that decoding the graphemic model with the 3-gram word LM does not give improvement.
\begin{table}[h]
	\centering
	\setlength{\tabcolsep}{0.5em}
	\vspace{-2mm}
	\caption{\it WERs (\%) for the \textbf{960hr} dataset.}
	\vspace{-4mm}
	\label{960overview}
	\begin{tabular}{ |l|c|c|c|c|c|} \hline
		\multirow{2}{*}{Unit} & \multirow{2}{*}{LM} & \multicolumn{2}{|c|}{dev} & \multicolumn{2}{|c|}{test}    \\ \cline{3-6}
		& & clean & other & clean & other    \\ \hline
		Phoneme      & 3-gram & 5.6 & 15.8 & 6.2 & 15.8  \\
		Grapheme     & None & 5.3 & 15.2 & 5.5 & 15.3  \\ 
		Word-Piece 16K & None & 4.4 & 13.2 & 4.7 & 13.4 \\ \hline
		Word-Piece 16K & LSTM & \textbf{3.3} & \textbf{10.3} & \textbf{3.6} & \textbf{10.3}  \\ \hline \hline
		\multirow{2}{*}{BPE 10K \cite{zeyer2018:asr-attention}} & None & 4.9 & 14.4 & 4.9 & 15.4 \\
		& LSTM & 3.5 & 11.5 & 3.8 & 12.8 \\ \hline
		\multirow{2}{*}{Hybrid system \cite{han2017capio}}  & N-gram & 3.4 & 8.8 & 3.6 & 8.9 \\
		& LSTM & 3.1 & 8.3 & 3.5 & 8.6 \\ \hline 
	\end{tabular}
	\vspace{-4mm}
\end{table}
\begin{table}[h]
	\centering
	\vspace{-2mm}
	\caption{\it Examples where the phonemic system's 1-best \textbf{wins} against the word-piece model's 1-best.}
	\vspace{-4mm}
	\label{plg_vs_wp}
	\begin{tabular}{ |c|c|} \hline
		Phoneme  & Word-Piece    \\ \hline
		when did you come \textbf{bartley} & when did you come partly    \\
		\textbf{kirkland} jumped for the jetty & kerklin jumped for the jetty \\ 
		man's eyes \textbf{remained} fixed & man's eyes were made fixed  \\ \hline
	\end{tabular}
		\vspace{-2mm}
\end{table}

In Table  \ref{960overview}, we also include the WERs from previous work on LibriSpeech 960hr; for fair comparison, systems which employ data augmentation \cite{park2019specaugment} are excluded.
Our word-piece model performs better than the previously reported sequence-to-sequence model
in \cite{zeyer2018:asr-attention} while the performance is behind the conventional hybrid system with
an n-gram LM \cite{han2017capio}. We note that our word-piece models simply trained using the cross-entropy criterion
(without e.g. minimum word error rate training \cite{PrabhavalkarSWN18}) is competitive with
Sabour et al.'s model trained with optimal completion distillation \cite{sabour2018optimal}, which is reported to give 4.5\% and 13.3\% on the test-clean and test-other sets.
For further comparison, we also report the WERs of our best word-piece model combined with an LSTM language model \cite{sundermeyer2012lstm} by shallow fusion \cite{ChorowskiJ17, toshniwal2018comparison}\footnote{In our experiments,
we find it crucial to constrain the emission of end-of-sentence (\texttt{EOS}) tokens \cite{ChorowskiJ17} to penalize short sentences
(rather than applying length normalization) in shallow fusion:
we only allow the model to emit \texttt{EOS} when its score is within 1.0 of the top hypothesis. We check that tuning such an \texttt{EOS} emission
constraint does not improve the baseline systems without language model
nor beam search with the WFST decoder for phonemic models.}.
The LSTM LM consists of one input linear layer of dimension 1024 and 2 LSTM layers with 2048 nodes \cite{hochreiter1997long}.
The LM weight of 0.35 is found to be optimal for dev-clean and dev-other WERs.
We obtain similar relative improvements reported in \cite{zeyer2018:asr-attention} and achieve WERs of 3.6\%
on the test-clean, and 10.3\% on the test-other set, which reduces the performance gap from the best hybrid system reported in \cite{han2017capio}.
\subsection{Results on 100hr and 460hr tasks}
\vspace{-2mm}
We conduct the same experiments in the 100hr and 460hr conditions.
For each unit, we obtain the best performance for the \textit{large} models
for the 460hr scenario, whereas for the 100hr case, the \textit{medium} model perform
the best. The results are summarized in Table \ref{460100overview}. We find that even
in the small dataset scenarios with higher OOV rates, graphemic and word-piece based
models outperform the phonemic system. We also note that the performance of attention-based
models dramatically degrades when the amount of training data is reduced, unlike conventional
hybrid approach \cite{panayotov2015librispeech}.
\begin{table}[h]
	\centering
	\vspace{-3mm}
	\caption{\it WERs for the \textbf{460hr} and \textbf{100hr} scenarios.}
		\vspace{-1mm}
				 	\vspace{-2mm}
	\label{460100overview}
	\begin{tabular}{ |c|c|c|c|c|c|c|} \hline
		Train   & \multirow{2}{*}{Unit} & \multicolumn{2}{|c|}{dev} & \multicolumn{2}{|c|}{test}    \\ \cline{3-6}
		data   &  & clean & other & clean & other    \\ \hline
		\multirow{3}{*}{460hr} & Phoneme&   7.6 &  27.3 & 8.5 & 27.8 \\
		& Grapheme&  6.4 & 23.5 & 6.8 &  24.1 \\ 
		& Word-Piece  & \textbf{5.7} & \textbf{21.8} & \textbf{6.5} & \textbf{22.5}  \\ \hline
		\multirow{3}{*}{100hr} &  Phoneme &  13.8 & 38.9 & 14.3 & 40.9 \\
		& Grapheme  &  \textbf{11.6} & 36.1 &  \textbf{12.0} &  38.0 \\ 
		& Word-Piece& 12.7 & \textbf{33.9} &  12.9 &  \textbf{35.5}  \\ \hline
		
	\end{tabular}
	\vspace{-5mm}
\end{table}
\vspace{-3mm}
\section{Rescoring Experiments}
\label{sec:part2}
\vspace{-2mm}
While Sec.~\ref{sec:part1} focuses on the comparison of models with different output
units, our goal is to ideally get benefits from different model units.
We consider two methods for combining LAS models with different output units.
The first approach is simple \textit{N-best list rescoring}.
We generate a N-best list from one LAS model, convert the corresponding word
sequences to the rescorer LAS model's unit, score them, and combine the scores
by log-linear interpolation to get new scores.
However, rescoring is limited to the hypotheses generated by one LAS model.
Therefore, we also carry out \textit{union of N-best list with cross-rescoring}:
we independently generate N-best lists from two LAS models, rescore the
hypotheses generated by one model using the other model and vice versa, to get the
1-best from the union of the rescored (up to) 2N hypotheses.
\vspace{-3mm}
\subsection{N-best Rescoring results}
\vspace{-2mm}
We carry out the N-best rescoring of our best word-piece based model in the 960hr scenario by a graphemic, and
a phonemic model. In all following experiments, the interpolation weights are optimized to obtain the best
dev-clean WER (which typically also gives the best dev-other WER).
The WERs are presented in the upper part of Table \ref{tab:resc_wp}.
We obtain improvements of 9\% in both cases on the test-clean set;
on the test-other set, we obtain an 8\% relative with the phonemic model and a 9\% relative with the graphemic model.
Thus, it can be noted that rescoring is a simple method for using
a phonemic model without an additional language model.
To determine if gains by the graphemic and phonemic models are additive, we
combine the scores from all models, which obtains only slight improvements of up
to 0.1 absolute as shown in Table \ref{tab:resc_wp} (+ Both).
In Table \ref{wp_gr_vs_wp_g_p}, we again show some illustrative examples
where the phonemic model outperforms the combination
of word-piece and grapheme based models only.
It is for example interesting to observe that the correct spelling ``bartley" is in the N-best hypotheses
of the word-piece model, and that the phonemic model helps recognize it correctly.

In the other direction, we also rescore the N-best list generated by the phonemic system by the word-piece model.
The results are shown in the lower part of Table \ref{tab:resc_wp}. We find that the improvements are limited (only
up to 4\% relative). In fact, the 30-best list generated by a phonemic system has much higher oracle WERs than the 8-best
list of the word-piece model.
\begin{table}[h]
	\centering
\setlength{\tabcolsep}{0.4em}
	\caption{\it WER (\%) results for N-best list rescoring. Oracle WERs are
	shown in parentheses.}
	\vspace{-4mm}
	\label{tab:resc_wp}
	\begin{tabular}{ |l|c|c|c|c|} \hline
		\multirow{2}{*}{ }  & \multicolumn{2}{|c|}{dev} & \multicolumn{2}{|c|}{test}    \\ \cline{2-5}
		& clean & other & clean & other    \\ \hline \hline
		Word-Piece        & 4.4 (2.4) & 13.2 (9.2) & 4.7 (2.6) & 13.4 (9.1) \\ 
		+ Phoneme         & 4.1  & 12.4 & \textbf{4.3} &  12.4 \\
		+ Grapheme        & 4.0 &  12.3 &  \textbf{4.3} &12.3 \\ 
		+ Both         & \textbf{3.9}  & \textbf{12.2} & \textbf{4.3} &  \textbf{12.2} \\ \hline \hline
		Phoneme  & 5.6 (4.9) & 15.8 (14.4) & 6.2 (5.5) & 15.8 (14.7) \\ 
		+ Word-Piece & 5.4 & 15.5 & 6.0 & 15.5 \\ \hline
	\end{tabular}
					 	\vspace{-3mm}
\end{table}
\begin{table}[h]
	\centering
	\caption{\it Examples where Word-Piece+Grapheme+Phoneme (WP+G+P) \textbf{wins} over Word-Piece+Grapheme (WP+G).}
				 	\vspace{-3mm}
	\label{wp_gr_vs_wp_g_p}
\resizebox{\columnwidth}{!}{
	\begin{tabular}{ |c|c|} \hline
		WP+G+\textbf{P}  & WP+G    \\ \hline
		oh \textbf{bartley} did you write to me & oh bartly did you write to me \\
	    ... lettuce leaf with \textbf{mayonnaise} ... & ... lettuce leaf with mayonna is ...    \\
       the manager \textbf{fell to} his musings & the manager felt of his musings \\
       what \textbf{a fuss} is made about you & what are fusses made about you \\
		... eyes \textbf{blazed with} indignation & ... eyes blaze of indignation  \\ \hline
	\end{tabular}
	  }
	\vspace{-5mm}
\end{table}
 \vspace{-3mm}
\subsection{Union of N-best lists with cross-rescoring results}
\vspace{-1mm}
The examples in Table \ref{plg_vs_wp} show some complementarity between the word-piece 1-best hypothesis
and the phonemic one.
To evaluate the potential value of hypotheses generated by the phonemic model, 
we decode a N-best list from the word-piece based and phoneme based models independently,
rescore the respective hypotheses (cross-rescoring), and take the 1-best from the 2N hypotheses (union).
In Table \ref{tab:union}, we observe that we only obtain marginal improvements on the test-other set, compared with rescoring the 8-best
word-piece hypotheses. For a fairer comparison, we also carry out rescoring of 16-best lists generated by
the word-piece model by the phonemic model. We find that such an approach is slightly better than the union.
This suggests that decoding from the phonemic model has limited benefits for the LibriSpeech task.
	\vspace{-1mm}
\begin{table}[h]
	\centering
	\setlength{\tabcolsep}{0.3em}
		\vspace{-2mm}
	\caption{\it WERs (\%) results for union of N-best lists with
		cross-rescoring. Oracle WERs are shown in parentheses.}
	\vspace{-3mm}
	\label{tab:union}
		\begin{tabular}{ |l|c|c|c|c|c|} \hline
			\multirow{2}{*}{ } & Num.  & \multicolumn{2}{|c|}{dev} & \multicolumn{2}{|c|}{test}    \\ \cline{3-6}
			& hyp.  & clean & other & clean & other    \\ \hline
			Word-Piece      & \multirow{2}{*}{8} & 4.4 (2.4) & 13.2 (9.2) & 4.7 (2.6) & 13.4 (9.1) \\
			+ Phoneme      &   & 4.1 & 12.4 & 4.3 & 12.4 \\ \cline{2-6}
			Union     & 16 & 4.1 & 12.4 & 4.3 & 12.3 \\ \hline
			Word-Piece     & \multirow{2}{*}{16} & 4.4 (2.0) & 13.2 (8.3) & 4.7 (2.2) & 13.4 (8.1)   \\ 
			+ Phoneme      &  & \textbf{4.0} & \textbf{12.3} & \textbf{4.3} &  \textbf{12.2}   \\ \hline

		\end{tabular}
\end{table}
 \vspace{-4mm}
\subsection{Why is Oracle WER So High for Phonemic System?}
      \vspace{-1mm}
The oracle WERs are much worse for the phonemic system than the word-piece model
(Table \ref{tab:resc_wp}).
We observe that the diversity of hypotheses in the N-best list generated by the phonemic
system is mainly based on homophones, rather than \textit{difficult} words (i.e. words with unusual
pronunciation).
For example, on the reference utterance \textit{``bozzle had always waited upon
	him with a decent coat and a well brushed hat and clean shoes"}, where
\textit{bozzle} is not in the training data, the word-piece based model fills the 8-best beam
by proposing different spellings for \textit{bozzle} such as \{basil, bazil,
basle, bosel, bosal, bosell, bossel\}, which is a reasonable way to model the ambiguity.
The phoneme system, instead, only produces \{bazil, basil\} as a substitution
for \textit{bozzle} and lists homophones for \textit{shoes}, \{shoes, shews,
shoos, shues, shooes\} instead.
Homophone distinction might still be inefficient for a phonemic system as the
phonemic LAS model gives them all the same score, and a single parameter is used
to weight the external LM for the entire search. 
Addressing this issue might be crucial to improve the phonemic system.
\subsection{Rescoring with an auxiliary decoder}
      \vspace{-1mm}
\begin{figure}[t]
  \centerline{\includegraphics[width=0.5\columnwidth]{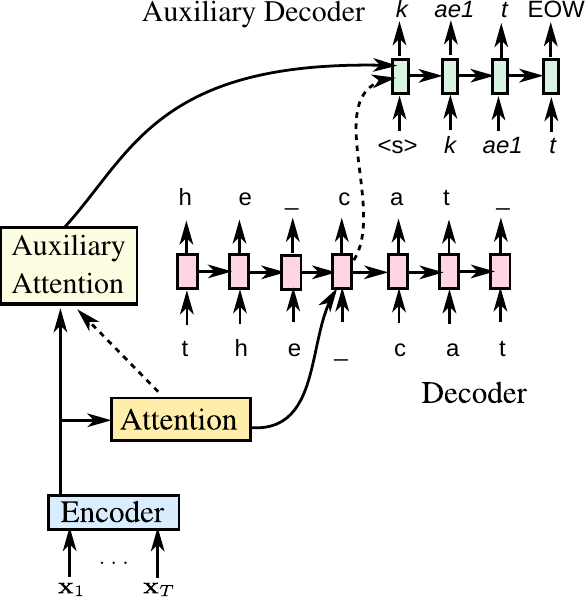}}
        \vspace{-1mm}
	\caption{\it LAS model with an auxiliary decoder: main
	decoder operates on graphemes and the auxiliary decoder predicts
	phonemes; dashed lines represent state copying for initialization at
	each word boundary.}
	\label{fig:las_aux}
	\vspace{-6mm}
\end{figure}
\label{sec:twodecmodel}
Finally, we examine a model with two decoders operating on different units but using a single encoder.
Such a model can be convenient for model combination (e.g., rescoring or potentially also decoding from two decoders operating on different units and combining hypotheses in a word synchronous fashion).
The design of the model is illustrated in Figure~\ref{fig:las_aux}.
The main decoder (grapheme, in the example) works exactly as in the baseline LAS model (Sec.~\ref{sec:model}; Figure~\ref{fig:las_base}).
The auxiliary decoder (phoneme, in the example) is designed such that it
predicts \emph{only the next word as a sequence of the auxiliary units}.
We use separate parameters for the auxiliary attention and initialize all recurrent states
of the auxiliary component at each word boundary by those of the main decoder (i.e. the prediction
from the auxiliary decoder is conditioned on the word sequence generated thus far from the main decoder).
The model is trained in two stages;
the main decoder and the encoder are first trained, and their parameters are not modified during the training of
the auxiliary components.
In experiments, we use word-pieces for the main decoder, and phonemes for the
auxiliary decoder.
Table \ref{tab:aux} shows improvements by rescoring with an auxiliary phoneme decoder of the two decoder-model.
We obtain improvements despite small number of additional parameters (30M) corresponding to the phonemic 2-layer LSTM decoder and the attention layer, however rescoring with an independent phoneme model (as in Table \ref{tab:resc_wp}) gives larger improvements.
\vspace{-3mm}
\begin{table}[h]
	\centering
	\setlength{\tabcolsep}{0.3em}
	\caption{\it WERs (\%) for rescoring with an auxiliary decoder.}
	\vspace{-3mm}
	\label{tab:aux}
	\begin{tabular}{ |l|c|c|c|c|l|} \hline
		\multirow{2}{*}{ }  & \multicolumn{2}{|c|}{dev} & \multicolumn{2}{|c|}{test}  & Total  \\ \cline{2-5}
		& clean & other & clean & other & Param.   \\ \hline
		Word-Piece (WP)           & 4.4 & 13.2 & 4.7 & 13.4  &180~M  \\
		WP + Auxiliary phoneme    &  4.3 & 13.0 &  4.6 &  13.1 &  210~M\\
		WP + Phoneme              &  4.1  & 12.4 &  4.3 &  12.4 & 310~M\\ \hline
		
	\end{tabular}
\end{table}
%
%
\vspace{-6mm}
\section{Conclusion}
Our experiments on different LibriSpeech subsets show that word-piece and grapheme based models
consistently outperform phoneme based models.
Therefore, the dominance of character-based model units in the LAS speech model
is not due to the amount of training data. This indicates that this behavior is more likely
related to the model itself (e.g., the decoder is conditioned on all predecessor labels).
Furthermore, we find that the word-piece based attention models can achieve a relatively low
oracle WER with only 8-best hypotheses and rescoring that N-best hypotheses using
graphemic or phonemic models gives good improvements.
Future work will examine whether streaming end-to-end approaches
(e.g., RNN-T \cite{graves2012sequence,RaoSP17}) show similar trends.
\label{sec:foot}
\clearpage

\bibliographystyle{IEEEtran}
\bibliography{main}

\begin{thebibliography}{10}
\providecommand{\url}[1]{#1}
\csname url@samestyle\endcsname
\providecommand{\newblock}{\relax}
\providecommand{\bibinfo}[2]{#2}
\providecommand{\BIBentrySTDinterwordspacing}{\spaceskip=0pt\relax}
\providecommand{\BIBentryALTinterwordstretchfactor}{4}
\providecommand{\BIBentryALTinterwordspacing}{\spaceskip=\fontdimen2\font plus
\BIBentryALTinterwordstretchfactor\fontdimen3\font minus
  \fontdimen4\font\relax}
\providecommand{\BIBforeignlanguage}[2]{{%
\expandafter\ifx\csname l@#1\endcsname\relax
\typeout{** WARNING: IEEEtran.bst: No hyphenation pattern has been}%
\typeout{** loaded for the language `#1'. Using the pattern for}%
\typeout{** the default language instead.}%
\else
\language=\csname l@#1\endcsname
\fi
#2}}
\providecommand{\BIBdecl}{\relax}
\BIBdecl

\bibitem{irie2019}
K.~Irie, R.~Prabhavalkar, A.~Kannan, A.~Bruguier, D.~Rybach, and P.~Nguyen,
  ``Model unit exploration for sequence-to-sequence speech recognition,''
  \emph{preprint arXiv:1902.01955}, 2019.

\bibitem{seq2seq}
I.~Sutskever, O.~Vinyals, and Q.~V. Le, ``Sequence to sequence learning with
  neural networks,'' in \emph{Advances in Neural Information Processing Systems
  (NIPS)}, Montreal, Canada, Dec. 2014, pp. 3104--3112.

\bibitem{bahdanau2014neural}
D.~Bahdanau, K.~Cho, and Y.~Bengio, ``Neural machine translation by jointly
  learning to align and translate,'' in \emph{Proc. Int. Conf. on Learning
  Representations (ICLR)}, San Diego, {CA}, {USA}, May 2015.

\bibitem{wu2016google}
Y.~Wu, M.~Schuster, Z.~Chen \emph{et~al.}, ``Google's neural machine
  translation system: Bridging the gap between human and machine translation,''
  \emph{arXiv preprint arXiv:1609.08144}, 2016.

\bibitem{RPis2017}
R.~Prabhavalkar, K.~Rao, T.~Sainath, B.~Li, L.~Johnson, and N.~Jaitly, ``A
  comparison of sequence-to-sequence models for speech recognition,'' in
  \emph{Proc. Interspeech}, Stockholm, Sweden, Aug. 2017, pp. 939--943.

\bibitem{KimHW17}
S.~Kim, T.~Hori, and S.~Watanabe, ``Joint ctc-attention based end-to-end speech
  recognition using multi-task learning,'' in \emph{Proc. {IEEE} Int. Conf. on
  Acoustics, Speech and Signal Processing (ICASSP)}, New Orleans, {LA}, {USA},
  Mar. 2017, pp. 4835--4839.

\bibitem{battenberg2017exploring}
E.~Battenberg, J.~Chen, R.~Child, A.~Coates, Y.~G.~Y. Li, H.~Liu, S.~Satheesh,
  A.~Sriram, and Z.~Zhu, ``Exploring neural transducers for end-to-end speech
  recognition,'' in \emph{Proc. {ASRU}}, Okinawa, Japan, Dec. 2017, pp.
  206--213.

\bibitem{WengCWWYSY18}
C.~Weng, J.~Cui, G.~Wang, J.~Wang, C.~Yu, D.~Su, and D.~Yu, ``Improving
  attention based sequence-to-sequence models for end-to-end english
  conversational speech recognition,'' in \emph{Proc. Interspeech}, Hyderabad,
  India, Sep. 2018, pp. 761--765.

\bibitem{ChiuSWPNCKWRGJL18}
C.~Chiu, T.~N. Sainath, Y.~Wu, R.~Prabhavalkar, P.~Nguyen, Z.~Chen, A.~Kannan,
  R.~J. Weiss, K.~Rao, E.~Gonina, N.~Jaitly, B.~Li, J.~Chorowski, and
  M.~Bacchiani, ``State-of-the-art speech recognition with sequence-to-sequence
  models,'' in \emph{Proc. {ICASSP}}, Calgary, Canada, Apr. 2018, pp.
  4774--4778.

\bibitem{sennrich16bpe}
R.~Sennrich, B.~Haddow, and A.~Birch, ``Neural machine translation of rare
  words with subword units,'' in \emph{ACL}, Berlin, Germany, August 2016, pp.
  1715--1725.

\bibitem{SchusterN12}
M.~Schuster and K.~Nakajima, ``Japanese and korean voice search,'' in
  \emph{Proc. {ICASSP}}, Kyoto, Japan, Mar. 2012, pp. 5149--5152.

\bibitem{lasicassp2016}
W.~Chan, N.~Jaitly, Q.~Le, and O.~Vinyals, ``Listen, attend and spell: a neural
  network for large vocabulary conversational speech recognition,'' in
  \emph{Proc. {ICASSP}}, Shanghai, China, Mar. 2016, pp. 4960--4964.

\bibitem{Bourlard1993}
H.~A. Bourlard and N.~Morgan, \emph{Connectionist Speech Recognition: A Hybrid
  Approach}.\hskip 1em plus 0.5em minus 0.4em\relax Norwell, {MA}, {USA}:
  Kluwer Academic Publishers, 1993.

\bibitem{kanthak02}
S.~Kanthak and H.~Ney, ``Context-dependent acoustic modeling using graphemes
  for large vocabulary speech recognition,'' in \emph{Proc. {IEEE} Int. Conf.
  on Acoustics, Speech and Signal Processing (ICASSP)}, Orlando, FL, USA, May
  2002, pp. 845--848.

\bibitem{KillerSS03}
M.~Killer, S.~St{\"{u}}ker, and T.~Schultz, ``Grapheme based speech
  recognition,'' in \emph{Proc. Eurospeech}, Geneva, Switzerland, Sep. 2003.

\bibitem{SungHBS09}
Y.~Sung, T.~Hughes, F.~Beaufays, and B.~Strope, ``Revisiting graphemes with
  increasing amounts of data,'' in \emph{{Proc. ICASSP}}, Taipei, Taiwan, Apr.
  2009, pp. 4449--4452.

\bibitem{sainath2017no}
T.~N. Sainath, R.~Prabhavalkar, S.~Kumar, S.~Lee, A.~Kannan, D.~Rybach,
  V.~Schogol, P.~Nguyen, B.~Li, and Y.~Wu, ``No need for a lexicon? evaluating
  the value of the pronunciation lexica in end-to-end models,'' in \emph{Proc.
  {ICASSP}}, Calgary, Canada, Apr. 2018, pp. 5859--5863.

\bibitem{zhou2018comparison}
S.~Zhou, L.~Dong, S.~Xu, and B.~Xu, ``A comparison of modeling units in
  sequence-to-sequence speech recognition with the {T}ransformer on {M}andarin
  {C}hinese,'' \emph{arXiv preprint arXiv:1805.06239}, 2018.

\bibitem{panayotov2015librispeech}
V.~Panayotov, G.~Chen, D.~Povey, and S.~Khudanpur, ``{LibriSpeech}: an {ASR}
  corpus based on public domain audio books,'' in \emph{{ICASSP}}, South
  Brisbane, Queensland, Australia, Apr. 2015, pp. 5206--5210.

\bibitem{WeissCJWC17}
R.~J. Weiss, J.~Chorowski, N.~Jaitly, Y.~Wu, and Z.~Chen,
  ``Sequence-to-sequence models can directly translate foreign speech,'' in
  \emph{Interspeech}, Stockholm, Sweden, Aug. 2017, pp. 2625--2629.

\bibitem{zeyer2018:asr-attention}
A.~Zeyer, K.~Irie, R.~Schl{\"u}ter, and H.~Ney, ``Improved training of
  end-to-end attention models for speech recognition,'' in \emph{Interspeech},
  Hyderabad, India, Sep. 2018, pp. 7--11.

\bibitem{bruguier19}
A.~Bruguier, R.~Prabhavalkar, G.~Pundak, and T.~N. Sainath, ``Phoebe:
  Pronunciation-aware contextualization for end-to-end speech recognition,'' in
  \emph{Proc. {ICASSP}}, Brighton, England, May 2019.

\bibitem{Brown92}
P.~F. Brown, P.~V. Desouza, R.~L. Mercer, V.~J.~D. Pietra, and J.~C. Lai,
  ``Class-based n-gram models of natural language,'' \emph{Computational
  linguistics}, vol.~18, no.~4, pp. 467--479, 1992.

\bibitem{mohri2008}
M.~Mohri, F.~Pereira, and M.~Riley, ``Speech recognition with weighted
  finite-state transducers,'' in \emph{Handbook of Speech Processing}.\hskip
  1em plus 0.5em minus 0.4em\relax Springer, 2008, ch.~28, pp. 559--582.

\bibitem{zhangCJ17}
Y.~Zhang, W.~Chan, and N.~Jaitly, ``Very deep convolutional networks for
  end-to-end speech recognition,'' in \emph{Proc. {IEEE} Int. Conf. on
  Acoustics, Speech and Signal Processing (ICASSP)}, New Orleans, {LA}, {USA},
  Mar. 2017, pp. 4845--4849.

\bibitem{schuster1997bidirectional}
M.~Schuster and K.~K. Paliwal, ``Bidirectional recurrent neural networks,''
  \emph{IEEE Transactions on Signal Processing}, vol.~45, no.~11, pp.
  2673--2681, 1997.

\bibitem{hochreiter1997long}
S.~Hochreiter and J.~Schmidhuber, ``Long short-term memory,'' \emph{Neural
  computation}, vol.~9, no.~8, pp. 1735--1780, 1997.

\bibitem{kingma15}
D.~P. Kingma and J.~Ba, ``Adam: A method for stochastic optimization,'' in
  \emph{ICLR}, San Diego, {CA}, {USA}, May 2015.

\bibitem{shen2019lingvo}
J.~Shen, P.~Nguyen, Y.~Wu, Z.~Chen \emph{et~al.}, ``Lingvo: a modular and
  scalable framework for sequence-to-sequence modeling,'' \emph{arXiv
  preprint:1902.08295}, 2019.

\bibitem{han2017capio}
K.~J. Han, A.~Chandrashekaran, J.~Kim, and I.~Lane, ``The {CAPIO} 2017
  conversational speech recognition system,'' \emph{arXiv preprint:1801.00059},
  2018.

\bibitem{park2019specaugment}
D.~S. Park, W.~Chan, Y.~Zhang, C.-C. Chiu, B.~Zoph, E.~D. Cubuk, and Q.~V. Le,
  ``Spec{A}ugment: A simple data augmentation method for automatic speech
  recognition,'' \emph{arXiv preprint arXiv:1904.08779}, 2019.

\bibitem{PrabhavalkarSWN18}
R.~Prabhavalkar, T.~N. Sainath, Y.~Wu, P.~Nguyen, Z.~Chen, C.~Chiu, and
  A.~Kannan, ``Minimum word error rate training for attention-based
  sequence-to-sequence models,'' in \emph{Proc. {ICASSP}}, Calgary, Canada,
  Apr. 2018, pp. 4839--4843.

\bibitem{sabour2018optimal}
S.~Sabour, W.~Chan, and M.~Norouzi, ``Optimal completion distillation for
  sequence learning,'' in \emph{Int. Conf. on Learning Representations (ICLR)},
  New Orleans, {LA}, {USA}, May 2019.

\bibitem{sundermeyer2012lstm}
M.~Sundermeyer, R.~Schl{\"u}ter, and H.~Ney, ``{LSTM} neural networks for
  language modeling.'' in \emph{Proc. Interspeech}, Portland, {OR}, {USA}, Sep.
  2012, pp. 194--197.

\bibitem{ChorowskiJ17}
J.~Chorowski and N.~Jaitly, ``Towards better decoding and language model
  integration in sequence to sequence models,'' in \emph{Internspeech},
  Stockholm, Sweden, Aug. 2017, pp. 523--527.

\bibitem{toshniwal2018comparison}
S.~Toshniwal, A.~Kannan, C.-C. Chiu, Y.~Wu, T.~N. Sainath, and K.~Livescu, ``A
  comparison of techniques for language model integration in encoder-decoder
  speech recognition,'' in \emph{Proc. {SLT}}, Athens, Greece, Dec. 2018.

\bibitem{graves2012sequence}
A.~Graves, ``Sequence transduction with recurrent neural networks,'' in
  \emph{Representation Learning Workshop, Int. Conf. on Machine Learning
  (ICML)}, Edinburgh, Scotland, Jun. 2012.

\bibitem{RaoSP17}
K.~Rao, H.~Sak, and R.~Prabhavalkar, ``Exploring architectures, data and units
  for streaming end-to-end speech recognition with rnn-transducer,'' in
  \emph{Proc. {ASRU}}, Okinawa, Japan, Dec. 2017, pp. 193--199.

\end{thebibliography}
\end{document}